\documentclass[sigconf]{acmart}

\renewcommand\footnotetextcopyrightpermission[1]{}

\AtBeginDocument{%
  }
\usepackage{tcolorbox}
\usepackage{lipsum}
\usepackage{multirow}
\usepackage{arydshln}
\usepackage{enumitem}
\usepackage{amssymb}
\usepackage{amsmath, amssymb}
\usepackage{graphicx}
\usepackage{adjustbox}
\usepackage{subcaption}
\usepackage{booktabs}
\usepackage{multirow}
\usepackage{tikz}
\usepackage{balance}
\usepackage[table]{xcolor}
\usepackage{tcolorbox}  
\definecolor{Mycolor1}{HTML}{A01D14}
\definecolor{Mycolor2}{HTML}{DBB15B}
\definecolor{Mycolor3}{HTML}{253D8C}
\definecolor{Mycolor4}{HTML}{256b48}

\newtcolorbox{mybox}{%
  breakable,enhanced,colback=white,colframe=black,left=0.5em,right=0.5em,boxrule=1.0pt}

\newtcolorbox{mybox3}[1]{
colbacktitle=white,coltitle=black,colback=white,colframe=black,fonttitle=\bfseries,title=#1,leftupper=0.5em,rightupper=0.5em,boxrule=0.6pt
}

\begin{document}

\title{ExDR: Explanation-driven Dynamic Retrieval Enhancement for Multimodal Fake News Detection}

\author{Guoxuan Ding}
\authornote{Equal Contribution. \{dingguoxuan,liyuqing\}@iie.ac.cn}
\affiliation{%
  \institution{Institute of Information Engineering, Chinese Academy of Sciences}%
  \country{China}%
}

\author{Yuqing Li}
\authornotemark[1]
\affiliation{%
  \institution{Institute of Information Engineering, Chinese Academy of Sciences}%
  \country{China}%
}

\author{Ziyan Zhou}
\affiliation{%
  \institution{Institute of Information Engineering, Chinese Academy of Sciences}%
  \country{China}%
}

\author{Zheng Lin}
\authornote{Zheng Lin is the corresponding author.}
\affiliation{%
  \institution{Institute of Information Engineering, Chinese Academy of Sciences}%
  \country{China}%
}

\author{Daren Zha}
\affiliation{%
  \institution{Institute of Information Engineering, Chinese Academy of Sciences}%
  \country{China}%
}

\author{Jiangnan Li}
\affiliation{%
  \institution{WeChat AI, Tencent}%
  \country{China}%
}

\renewcommand{\shortauthors}{Li et al.}

\begin{abstract}
The rapid spread of multimodal fake news poses a serious societal threat, as its evolving nature and reliance on timely factual details challenge existing detection methods. 
Dynamic Retrieval-Augmented Generation provides a promising solution by triggering keyword-based retrieval and incorporating external knowledge, thus enabling both efficient and accurate evidence selection. However, it still faces challenges in addressing issues such as redundant retrieval, coarse similarity, and irrelevant evidence when applied to deceptive content.
In this paper, we propose ExDR—an Explanation-driven Dynamic Retrieval-Augmented Generation framework for Multimodal Fake News Detection. Our framework systematically leverages model-generated explanations in both the retrieval triggering and evidence retrieval modules. It assesses triggering confidence from three complementary dimensions, constructs entity-aware indices by fusing deceptive entities, and retrieves contrastive evidence based on deception-specific features to challenge the initial claim and enhance the final prediction.
Experiments on two benchmark datasets, AMG and $\text{MR}^2$, demonstrate that ExDR consistently outperforms previous methods in retrieval triggering accuracy, retrieval quality, and overall detection performance, highlighting its effectiveness and generalization capability.
\end{abstract}

\ccsdesc[300]{Information systems~Multimedia information systems}
\ccsdesc[500]{Computing methodologies~Natural language generation}

\keywords{Fake News Detection, Multimodel Analysis, Dynamic Retrival Method}

\maketitle

\section{Introduction}
The rapid spread of social media has dramatically accelerated the spread of fake news~\cite{shu2017fake}, especially in multimodal formats combining text and images. Compared to traditional text-based misinformation, multimodal fake news leverages the stronger emotional appeal and vividness of visual content to achieve faster dissemination and broader societal impact. Detecting multimodal fake news has thus emerged as a critical and urgent challenge in the field of information security.

Recent Multimodal Fake News Detection (MFND) studies mainly rely on the combination of multiple language and vision models, designing complex and structured architectures to extract and align features from textual and visual modalities~\cite{dong2024unveiling, tong2024mmdfnd}, facing challenges in cross-modal feature fusion and requiring aligned multimodal training data.
Large Vision-Language Models (LVLMs), with their unified image-text embedding space and rich world knowledge, demonstrate strong adaptability to multimodal tasks~\cite{liu2024fka}. 
However, the reliance on internal memory accumulated during training reduces its effectiveness in addressing newly emerging knowledge or long-tail instances, often resulting in hallucinations.

Dynamic Retrieval-Augmented Generation (RAG) offers a promising alternative by triggering keyword-based retrieval and incorporating external information retrieval instead of over-relying on the model’s inherent reasoning capabilities~\cite{abootorabi2025ask,mei2025survey}. This contrasts with static RAG~\cite{sohn2025rationale,wu2025sitembv15improvedcontextawaredense,li2025mindscapeawareretrievalaugmentedgeneration} pipelines that retrieve for every input regardless of necessity. With a decoupled retriever that can be updated independently of the model, dynamic RAG is well-suited for LVLMs to handle evolving knowledge and long-tail cases.
However, naive application of dynamic RAG introduces three key issues. 
First, for straightforward or commonsense-based fake news, LVLMs often already possess sufficient knowledge. Blindly triggering retrieval without considering sample difficulty leads to unnecessary noise and computational overhead~\cite{su2024dragin,jeong2024adaptive}.
Second, most existing retrieval strategies focus on sample-level similarity while neglecting finer-grained internal cues~\cite{tang2024leveraging,mei2025survey}, which are often critical for accurately identifying deceptive content.
Finally, retrieved results tend to include only the most similar samples, which often lack deception-specific signals and may introduce useless or misleading evidence, ultimately impairing fake news detection accuracy~\cite{tang2024leveraging}.

To address these challenges, we propose a novel framework named \textbf{ExDR}: \textbf{Ex}planation-driven \textbf{D}ynamic \textbf{R}etrieval, specifically designed for Multimodal Fake News Detection.
Our framework consists of two main modules: the retrieval triggering and the evidence retrieval, both of which leverage an explanation-driven RAG paradigm to better adapt to the requirements of MFND. ExDR places model-generated explanations at the center of the retrieval process and systematically investigates three key questions: when to trigger retrieval, how to construct effective retrieval queries, and what types of deceptive samples to retrieve. Each of these aspects targets a distinct challenge inherent in RAG-based methods, enabling more accurate and efficient detection of multimodal deception.

\textbf{For retrieval triggering}, we design a dynamic triggering strategy tailored for MFND. This strategy constructs three confidence dimensions derived from the model-generated explanation, enabling a comprehensive assessment of whether retrieval is necessary for each prediction.  To determine the optimal triggering thresholds for these indicators, we develop a two-stage hybrid search strategy that efficiently explores the threshold space. Notably, prior dynamic RAG methods typically iterate retrieval to refine the retrieval target, whereas we adopt a one-shot scheme: a single trigger decision followed by at most one retrieval step per instance, with the retrieved evidence directly used for the final prediction for efficiency.

\textbf{For evidence retrieval}, we build an entity-enriched multimodal hybrid indexing strategy and perform contrastive evidence retrieval guided by model-generated
explanations. This process integrates visual, textual, and explanation-derived entity information into the index, enabling more deception-relevant and semantically aligned retrieval. Subsequently, based on the inferred fine-grained deception labels, we retrieve both positive and negative evidence to provide the model with more targeted guidance for model decision-making.

To comprehensively evaluate dynamic retrieval strategies in our task, we further introduce two metrics: \textbf{Retrieval Identification Rate}, which measures the accuracy of recognizing when retrieval is needed, and \textbf{Retrieval Efficiency}, which quantifies the performance gains attributable to retrieval triggering.

We conduct systematic evaluations of the proposed framework on two widely used MFND datasets, AMG and MR2. Experimental results demonstrate that our method significantly outperforms existing approaches in retrieval triggering accuracy, retrieval quality, and final detection performance, highlighting its strong practicality and generalization capabilities. 

In summary, our main contributions are as follows:

\begin{itemize}

\item We make the first attempt to apply the dynamic Retrieval-Augmented Generation framework to Multimodal Fake News Detection. Differing from prior supervised methods, our approach enables efficient detection without the need for task-specific training.
\item We propose ExDR, an explanation-driven framework that innovates on retrieval triggering and contrastive evidence retrieval, achieving more efficient and deception-targeted retrieval.

\item We design two new metrics to assess the accuracy and necessity of retrieval triggering. Extensive experiments on both in-domain and out-of-domain datasets demonstrate that ExDR achieves state-of-the-art detection performance, while our new metrics confirm its superiority in retrieval precision and necessity.
\end{itemize}
\section{Related Work}

The goal of multimodal fake news detection is to identify inconsistencies or deceptive content across modalities such as text and images. BMR~\cite{ying2023bootstrapping} employ single-view prediction to disentangle multimodal features and explain which modality plays a critical role in the final decision. NSLM~\cite{dong2024unveiling} leverages symbolic logic constraints to unveil deceptive patterns within multimodal fake news. MMDFND~\cite{tong2024mmdfnd} enhances multimodal feature representation by integrating additional vision and language models and applying feature-wise weighting strategies. FKA-Owl~\cite{liu2024fka} encodes forgery-specific features as learnable continuous vectors and utilizes soft prompt tuning to strengthen the reasoning ability of LVLMs. Although these approaches improve the effectiveness of multimodal fake news detection, they remain limited by challenges in modality alignment and a strong reliance on high-quality deception-specific features. Retrieval-Augmented Generation (RAG) alleviates these issues by incorporating relevant information retrieved from external corpora~\cite{rag1,rag2,rag3,rag4,rag5,rag6}. More recently, dynamic RAG techniques further enhance this capability by adaptively deciding when to retrieve external evidence and what to retrieve during inference, thereby improving model performance and task accuracy~\cite{drag1,drag3,drag4,drag5}. Building upon this line of work, we introduce an improved dynamic RAG framework that enables accurate and efficient identification of deceptive cues, leading to more effective and efficient multimodal fake news detection.

\section{Method}

\begin{figure*}[htb]
\centering
\includegraphics[width=0.99\textwidth]{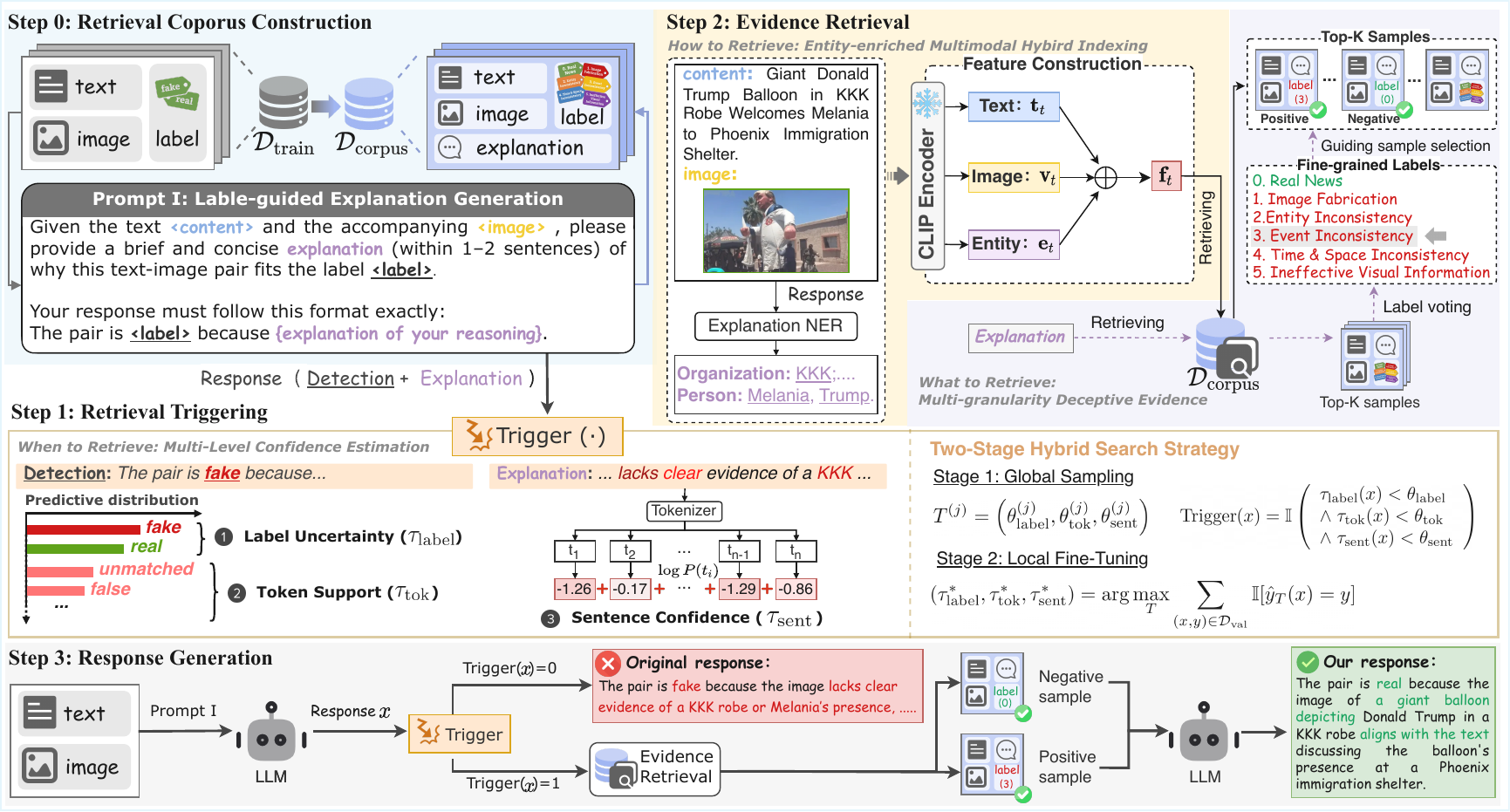}
\caption{Overview of our proposed ExDR framework. ExDR consists of two main components: (1) a retrieval triggering module that dynamically determines whether retrieval is necessary based on response analysis, and (2) an evidence retrieval module that retrieves targeted evidence, including both positive and negative samples guided by fine-grained deception labels, to enrich the context and improve model generation.}
\label{fig1}
\end{figure*}
In this section, we introduce \textbf{ExDR}, a retrieval-augmented framework for Multimodal Fake News Detection, as illustrated in Figure~\ref{fig1}. ExDR consists of two core modules. The retrieval triggering module leverages confidence estimation derived from the model's output to decide when retrieval is necessary. 
The evidence retrieval module fetches targeted supporting evidence using entity-enriched features and multi-granularity deception signals. Together, these modules jointly enhance retrieval efficiency and detection performance.

\subsection{Preliminaries}
\label{sec:pre}

\subsubsection{Task Definition}

A Multimodal Fake News Detection (MFND) dataset is defined as $\mathcal{D} = \{(I_i, T_i, y_i)\}$, where $I_i$ represents an image, $T_i$ denotes the textual content, $y_i \in \{\textsc{Real}, \textsc{Fake}\} $ is the binary label for the pair $(I_i, T_i)$.
\subsubsection{Data Preparation}
We construct our retrieval corpus from the official training split of the AMG~\cite{guo2025each} dataset. This dataset provides not only binary labels but also fine-grained deception labels $\tilde{y}_i$ for each fake news sample, as shown in \textsc{Appendix}~\ref{appendix:deception_label}. To enrich this corpus with explicit reasoning, we leverage GPT-4o to generate a textual explanation $E_i$ for each sample $(I_i, T_i)$, justifying its corresponding label $\tilde{y}_i$. The final retrieval corpus is thus a collection of tuples $D_{corpus} = \{(I_i, T_i, E_i, \tilde{y}_i)\}$. The prompt used for explanation generation is shown in Figure~\ref{fig1}, and the quality of the generated explanations is validated via human evaluation in Appendix~\ref{appendix:human_verify}.

Moreover, the AMG validation set $\mathcal{D}_{\text{val}}$ is used to determine the optimal configuration for retrieval triggering.

\subsubsection{Response Formation}
Unlike traditional Multimodal Fake News Detection methods that typically output only a classification label, we adopt a Chain-of-Thought~\cite{wei2022chain} prompting format to generate both the answer and the reasoning process from the model. Specifically, the model's response consists of two parts:

\textbf{1) Detection}
determines whether the input image--text pair is labeled as \textsc{Real} or \textsc{Fake}.
\textbf{2) Explanation}
provides the rationale behind the detection decision.

Each part serves as an independent retrieval trigger, and the final retrieval decision is made based on their joint signals.
Detailed prompt used for response generation is provided in Appendix~\ref{sec:appendix_prompting}. 
\subsection{When to Retrieve: Multi-Level Confidence Estimation}
\label{sec:when}
Most dynamic RAG works, which are typically designed for generation tasks such as QA, trigger retrieval based on next-token uncertainty~\cite{su2024dragin,jiang2023active}.
This token-centric approach is not optimal for classification, where the final prediction is a single label rather than a token sequence.
To bridge this gap, we introduce a novel triggering mechanism that evaluates the model's confidence based on its decision explanation. Specifically, we design a multi-level confidence evaluation method that integrates three complementary dimensions centered on the predicted label: Label-level Uncertainty, Token-level Support, and Sentence-level Confidence.

\subsubsection{Label-level Uncertainty.}
Label-level Uncertainty serves as a fundamental and explicit indicator of the model’s certainty in assigning a predefined label. It provides a direct measure of the reliability of the classification decision. Formally, we define it as:
\begin{equation}
     \tau_{\text{label}} = \left| \frac{\log p_{\text{real}} - \log p_{\text{fake}}}{\log p_{\text{real}} + \log p_{\text{fake}}} \right| ,
\end{equation}
where $p_{\text{real}}$ and $p_{\text{fake}}$ denote the generation probabilities for the \textsc{Real} and \textsc{Fake} labels, respectively, obtained from the model under the prompt-based classification setting.

\subsubsection{Token-level Support.}
While explicit classification indicators such as \textsc{Real} and \textsc{Fake} provide valuable signals, they may not fully reflect the model's decision-making process, particularly when the model exhibits overconfidence or when its predictions are influenced by localized biases. To obtain a more comprehensive understanding of the model's confidence, we analyze the distribution of the top-$K$ candidate tokens at the classification position.

We evaluate each candidate token's semantic alignment with the predicted label through hierarchical matching: direct lexical matching against curated lexicons $\mathcal{L}_{\text{real}}$ and $\mathcal{L}_{\text{fake}}$ (e.g., ``genuine", ``false"), followed by semantic similarity matching for out-of-lexicons tokens. The token-level support score is then computed as:

\begin{equation} \tau_{\text{tok}} = \frac{N_{\text{sup}}}{K}, \end{equation}
where $N_{\text{sup}}$ denotes the count of top-$K$ candidate tokens that semantically support the model's current prediction.
The detail of $\mathcal{L}_{\text{real}}$ and $\mathcal{L}_{\text{fake}}$, along with the alignment method, is provided in the \textsc{Appendix}~\ref{appendix:Lexicons}.
\subsubsection{Sentence-level Confidence. }
In addition to token-level analysis, we compute the perplexity of the model-generated explanation as an indicator of the model’s reasoning-level confidence toward the predicted label. The sentence-level confidence is defined as:
\begin{equation}
\tau_{\text{sent}} = \exp\left(\frac{1}{n} \sum_{i=1}^{n} \log P(t_i)\right) ,
\end{equation}
where $P(t_i)$ denotes the generation probability of the $i$-th token in the explanation, and 
$n$ is the total number of tokens. 

Centered on the predicted class label, these three indicators collectively evaluate the model’s output across multiple dimensions, including the label logit, the distribution of candidate tokens, and the coherence of the generated reasoning.

\subsubsection{Confidence Threshold Optimization}
After establishing the evaluation indicators, we determine thresholds for triggering retrieval. Our goal is to improve MFND by selectively activating retrieval based on the three confidence scores. As the thresholds are continuous, exhaustive grid search is computationally prohibitive. We therefore adopt a two-stage hybrid strategy that balances \textbf{global exploration} and \textbf{local exploitation}.

We first conduct a coarse-grained global exploration scan over the entire feasible threshold space $\mathcal{T}$, where each dimension corresponds to the empirical range of confidence scores observed on $\mathcal{D}{\text{val}}$. Specifically, given $N_{iter}$ sampling iterations, we perform Monte Carlo sampling to draw candidate threshold triplets:

\begin{equation}
T^{(j)} = \left( \theta^{(j)}_{\text{label}}, \theta^{(j)}_{\text{tok}} ,\theta^{(j)}_{\text{sent}} \right), \quad j = 1, 2, \ldots, N_{iter}.
\end{equation}

For each combination $T^{(j)}$, we define the retrieval triggering logic on $\mathcal{D}_{\text{val}}$:
\begin{equation}
\begin{split}
\text{Trigger}^{(j)}(x) = \mathbb{I} \big[ \tau_{\text{label}}(x) < \theta^{(j)}_{\text{label}} &\land \tau_{\text{tok}}(x) < \theta^{(j)}_{\text{tok}}\\
&\land \tau_{\text{sent}}(x) < \theta^{(j)}_{\text{sent}} \big] ,
\end{split}
\end{equation}
where $\mathbb{I}\left[\cdot \right]$ is the indicator function. 

Once retrieval is triggered, the retrieved samples are incorporated into the prompt as context, and the model generates a new prediction based on this augmented input.

The performance of each combination is evaluated by: 
\begin{equation} 
\text{Score}^{(j)} = \sum_{(x, y) \in \mathcal{D}_{\text{val}}} \mathbb{I}[ \hat{y}_{T^{(j)}}(x) = y ],
\end{equation}
where $\hat{y}_{{T}^{(x)}}$ denotes the final prediction, selected between the original and retrieval-augmented predictions according to the $\text{Trigger}(x)$.

After the global exploration phase, we identify the top-$K$ triplets of threshold configurations that yield the best performance. These top-performing configurations serve as central points for subsequent fine-grained local exploitation.
For each of the $K$ selected centers, we perform a localized grid search in their immediate neighborhood by perturbing each threshold dimension with a small step size $\delta$. The final optimal threshold configuration is selected as the one that achieves the highest accuracy on $D_{val}$ across all candidates evaluated during both the global and local search stages:
\begin{equation}
(\theta_{\text{label}}^*, \theta_{\text{tok}}^*, \theta_{\text{sent}}^*) = \arg\max_{T} \text{Score}(T) .
\end{equation}

By combining global and local search, this approach ensures both efficient exploration and precise tuning, leading to a well-generalized threshold configuration, which is subsequently fixed and used for experiments.

\subsection{How to Retrieve: Entity-enriched Multimodal Hybrid Indexing}
After retrieval is triggered, we construct an effective indexing strategy based on key entities derived from the news content. 
Prior multimodal approaches typically rely solely on image-text representations~\cite{RGCL2024Mei}, which often contain noisy or irrelevant information and overlook entities that are critical for identifying deception.
To address this limitation, we propose an explanation-guided entity-enriched multimodal hybrid indexing method. We extract key entities from the model-generated explanation, as they tend to highlight potentially deceptive content. 
These entities are more likely to reflect inconsistencies or fabrications, making them effective anchors for constructing semantically meaningful indexing representations.

Specifically, we begin by employing a pre-trained Named Entity Recognition (NER) model to extract a set of key entities $\mathcal{E}_t = \{e_1, \dots, e_t, \dots\}$ from the explanation $E_t$. 
These entities serve as potential deception indicators.

Next, we leverage a frozen CLIP encoder to obtain three hybrid feature vectors: (1) the visual feature $\mathbf{v}_t$ extracted from the image, (2) the textual feature $\mathbf{t}_t$ derived from the original text, and (3) the entity feature $\mathbf{e}_t$ obtained by encoding a natural-language concatenation of the extracted entities. To integrate these modalities, we compute an averaged feature:

\begin{equation}
\mathbf{f}_t = \frac{ \mathbf{v}_t + \mathbf{t}_t + \mathbf{e}_t }{3} ,
\end{equation}
where $\mathbf{f}_t$ jointly captures the visual context, the textual claim, and the explanation-guided entity information. 

The fused feature $\mathbf{f}_t$ is subsequently $L_2$-normalized to yield the final entity-enriched indexing representation:

\begin{equation}\quad 
\hat{\mathbf{f}}_t = \frac{\mathbf{f}_t}{\|\mathbf{f}_t\|_2}.
\end{equation}

By enriching image-text features with entity information, this entity-enriched indexing facilitates more targeted and semantically aligned retrieval for MFND.

\subsection{What to Retrieve: Multi-Granularity Deceptive Evidence}
Standard evidence retrievals often retrieve broadly similar samples, overlooking the fine-grained, deception-specific evidence required to verify a claim.
We address this through a two-step process: 
(1) infer a fine-grained deception label from the model’s explanation, and (2) use the inferred label to retrieve targeted contrastive examples.

\subsubsection{Explanation-Guided Label Inference}
We assign a fine-grained deception label $\hat{y}_t$ to the model-generated explanation $E_t$.
First, we compute the cosine similarity between $E_t$ and  explanations $E_i$ in the corpus $D_{\text{corpus}}$, retrieving the top-$K$ most similar explanations and their corresponding fine-grained labels $\tilde{y}_i$:

\begin{equation}
N_t = \operatorname{Top}K \left\{ \cos(E_t, E_i) \mid i \in D_{\text{corpus}} \right\},
\end{equation}
where $N_t$ denotes the set of indices of the top-$K$ most similar explanations, and $\cos(\cdot, \cdot)$ denotes the cosine similarity.

Next, the infered fine-grained label $\hat{y}_t$ is determined via a majority vote among the fine-grained labels $\tilde{\mathbf{y}}$ of the retrieved candidates:
\begin{equation}
\hat{{y}}_t = \underset{c\in \mathbf{\tilde{y}}}{\arg\max} \sum_{i \in N_t} \mathbb{I}(\tilde{y}_i = c).
\end{equation}

\subsubsection{Contrastive Evidence Retrieval}
With the entity-enriched feature $\hat{\mathbf{f}}_t$ and the inferred fine-grained label $\hat{y}_t$, we retrieve informative contrastive evidence to support or refute the model's prediction. 
To enable efficient similarity search, we build an FAISS index over the fused features ${\hat{\mathbf{f}}_i} \in D_{\text{corpus}}$, and adopt the dot product as the similarity metric. We retrieve two types of contrastive examples: a positive instance $\mathbf{x}^{+}$ whose fine-grained label matches $\hat{y}_t$, and a negative instance $\mathbf{x}^{-}$ whose binary label differs from $y_t$:

\begin{equation} s_i = \hat{\mathbf{f}}_t \cdot \hat{\mathbf{f}}_i, \end{equation}

\begin{equation} \mathbf{x}^{+} = \underset{i: \hat{{y}}_t = \tilde{{y}}_i}{\arg\max} \: s_i, \end{equation}

\begin{equation} \mathbf{x}^{-} = \underset{i: {y}_t \neq {y}_i, }{\arg\max} \: s_i.\end{equation}

Finally, the two retrieved samples are prepended to the original input to form the full context $[\mathbf{x}^{+};\mathbf{x}^{-};\mathbf{x}]$. This contrastive context exposes the model to both supporting and opposing instances, facilitating a more informed decision under ambiguous or hard cases. Prompt details are shown in Appendix~\ref{sec:appendix_prompting}.

\section{Experiments}

\begin{table*}[t]
\centering
\caption{Results of different retrieval triggering methods on the \texttt{AMG} and \texttt{MR$^2$} datasets. For each model, we report RI and RE scores for both the \colorbox{lightgray}{\textit{vanilla}} and \textbf{fine-tuned} versions. * indicates that no retrieval is triggered for that method.}
\begin{adjustbox}{width=\textwidth}
\begin{tabular}{l|cccc|cccc|cccc|cccc|cccc}
\toprule
\textbf{Methods} 
& \multicolumn{4}{c}{\textbf{Qwen-2.5-VL-7b}} 
& \multicolumn{4}{c}{\textbf{InternVL2.5-8b}} 
& \multicolumn{4}{c}{\textbf{LLaVA-1.6-Mistral-7b}} 
& \multicolumn{4}{c}{\textbf{Qwen-2.5-VL-32b}} 
& \multicolumn{4}{c}{\textbf{InternVL2.5-26b}} \\

& \multicolumn{2}{>{\columncolor{gray!15}}c}{\textit{RI} \quad \textit{RE}} 
& \textbf{RI} & \textbf{RE} 
& \multicolumn{2}{>{\columncolor{gray!15}}c}{\textit{RI} \quad \textit{RE}} 
& \textbf{RI} & \textbf{RE} 
& \multicolumn{2}{>{\columncolor{gray!15}}c}{\textit{RI} \quad \textit{RE}} 
& \textbf{RI} & \textbf{RE} 
& \multicolumn{2}{>{\columncolor{gray!15}}c}{\textit{RI} \quad \textit{RE}} 
& \textbf{RI} & \textbf{RE} 
& \multicolumn{2}{>{\columncolor{gray!15}}c}{\textit{RI} \quad \textit{RE}} 
& \textbf{RI} & \textbf{RE} \\
\midrule
\midrule

\multicolumn{9}{l}{\textit{AMG (in-domain)}} \\

FL-RAG
& \textbf{41.5} & 0.91 & \underline{16.7} & \underline{1.19}
& \underline{41.6} & \underline{1.14} & 18.0 & -0.36
& 47.8 & \underline{1.85} & \underline{14.1} & 0.00
& \textbf{38.9} & 0.70 & * & *
& \textbf{50.5} & \textbf{3.75} & * & * \\

FLARE
& 38.3 & \textbf{1.34} & 15.2 & -8.45
& 39.4 & 1.01 & * & *
& \underline{49.9} & 1.74 & * & *
& 30.8 & \textbf{1.01} & \underline{18.6} & \underline{2.47}
& 32.6 & 1.20 & * & * \\

DRAGIN
& 36.1 & 0.98 & * & *
& 39.5 & 1.00 & \underline{16.6} & \underline{3.00}
& 49.7 & 1.72 & 12.4 & \underline{1.00}
& \underline{31.0} & \underline{0.88} & 14.5 & 0.00
& 29.8 & \underline{2.13} & \underline{11.7} & \underline{0.00} \\

\textbf{ExDR}$_{\textit{trigger}}$
& \underline{38.8} & \underline{1.32} & \textbf{50.0} & \textbf{15.48}
& \textbf{43.8} & \textbf{1.22} & \textbf{60.0} & \textbf{33.03}
& \textbf{64.3} & \textbf{2.55} & \textbf{66.7} & \textbf{+28.15}
& 30.1 & 0.87 & \textbf{52.1} & \textbf{23.33}
& \underline{37.0} & 1.62 & \textbf{58.5} & \textbf{11.33} \\

\midrule

\multicolumn{9}{l}{\textit{MR$^2$ (cross-domain)}} \\

FL-RAG
& 37.5 & 1.79 & 26.7 & \textbf{0.87}
& \underline{42.0} & \underline{1.07} & \underline{34.4} & \underline{22.50}
& 49.0 & \underline{1.39} & \underline{40.2} & \underline{1.32}
& 29.3 & \underline{2.17} & * & *
& \textbf{38.2} & \textbf{+4.35} & * & * \\

FLARE
& \textbf{40.3} & \textbf{3.46} & \underline{29.4} & 0.00
& 41.4 & 1.02 & * & *
& \underline{49.7} & 1.25 & * & *
& \underline{33.9} & 1.78 & \underline{34.5} & \textbf{2.42}
& 31.4 & 2.39 & * & * \\

DRAGIN
& 37.1 & 1.01 & 0.0 & 0.00
& 41.7 & 1.02 & 33.2 & 12.60
& 49.4 & 1.25 & 35.3 & 0.98
& 33.0 & 1.00 & 28.7 & \underline{1.29}
& \underline{33.5} & 0.67 & \underline{29.9} & \underline{+1.03} \\

\textbf{ExDR}$_{\textit{trigger}}$
& \underline{37.9} & \underline{2.14} & \textbf{31.4} & -1.13
& \textbf{42.9} & \textbf{1.37} & \textbf{53.4} & \textbf{51.91}
& \textbf{57.6} & \textbf{1.73} & \textbf{68.7} & \textbf{2.54}
& \textbf{34.7} & \textbf{2.69} & \textbf{36.8} & 0.60
& 29.9 & \underline{8.07} & \textbf{48.5} & \textbf{+4.63} \\

\bottomrule
\end{tabular}

\end{adjustbox}

\label{tab:trigger-results}
\end{table*}

\begin{table*}[t]
\centering
\caption{Results of different evidence retrieval methods on the \texttt{AMG} and \texttt{MR$^2$} datasets. We report Accuracy (ACC) and F1 scores for both the \colorbox{lightgray}{\textit{vanilla}} and \cellcolor{lightgray}\textbf{fine-tuned} versions. Both the @full and @dynamic settings utilize our proposed explanation-guided contrastive evidence retrieval, where @full applies it to every sample and @dynamic employs our retrieval trigger component.}
\begin{adjustbox}{width=\textwidth}
\begin{tabular}{l|cccc|cccc|cccc|cccc|cccc}
\toprule
\textbf{Methods} 
& \multicolumn{4}{c}{\textbf{Qwen-2.5-VL-7b}} 
& \multicolumn{4}{c}{\textbf{InternVL2.5-8b}} 
& \multicolumn{4}{c}{\textbf{LLaVA-1.6-Mistral-7b}} 
& \multicolumn{4}{c}{\textbf{Qwen-2.5-VL-32b}} 
& \multicolumn{4}{c}{\textbf{InternVL2.5-26b}} \\

& \multicolumn{2}{>{\columncolor{gray!15}}c}{\textit{ACC} \quad \textit{F1}} 
& \textbf{ACC} & \textbf{F1} 
& \multicolumn{2}{>{\columncolor{gray!15}}c}{\textit{ACC} \quad \textit{F1}} 
& \textbf{ACC} & \textbf{F1} 
& \multicolumn{2}{>{\columncolor{gray!15}}c}{\textit{ACC} \quad \textit{F1}} 
& \textbf{ACC} & \textbf{F1} 
& \multicolumn{2}{>{\columncolor{gray!15}}c}{\textit{ACC} \quad \textit{F1}} 
& \textbf{ACC} & \textbf{F1} 
& \multicolumn{2}{>{\columncolor{gray!15}}c}{\textit{ACC} \quad \textit{F1}} 
& \textbf{ACC} & \textbf{F1} \\
\midrule
\midrule

\multicolumn{9}{l}{\textit{AMG (in-domain)}} \\

wo-RAG
& 63.5 & 67.5 & 84.2 & 83.2 & 60.4 & 64.3&84.5&83.3&59.1&65.3&85.9&84.9&69.0&71.3&86.0&84.8&66.8&64.8&85.3&83.9 \\

Text@full
&61.8&64.7&84.8&83.4&52.4&61.0&84.5&83.3&65.9&66.2&85.5&84.5&65.1&68.4&86.4&85.1&58.0&62.3&85.0&83.9 \\

Text+Image@full
&\textbf{73.8}&\textbf{71.6}&85.3&82.8&66.8&64.4&85.1&82.5&58.1&55.0&84.6&83.7&71.5&71.5&\underline{87.1}&85.0&69.1&63.4&\textbf{87.5}&\textbf{85.8} \\

\hline
\textbf{ExDR}$_\textit{evidence}$& & & & & & & & & & & & & & & & & & & &
 \\

\hspace{4mm}-single@full
&70.4&\underline{71.3}&\underline{85.4}&83.5&\textbf{71.2}&\textbf{68.4}&85.5&83.3&71.5&54.5&84.7&86.7&\textbf{73.5}&\underline{72.8}&83.7&85.1&\underline{69.3}&\textbf{67.3}&86.1&\underline{84.6} \\

\hspace{4mm}-single@dynamic
&69.6&70.8&85.0&\underline{83.7}&68.1&\underline{66.7}&85.4&\underline{83.9}&67.7&\underline{69.7}&\underline{86.1}&\underline{85.1}&71.2&72.2&\textbf{87.2}&\textbf{85.7}&67.3&65.2&85.7&84.3 \\
\hspace{4mm}-pos+neg@full
&\underline{71.8}&70.4&\textbf{86.0}&\textbf{83.8}&\underline{69.2}&66.2&\underline{85.8}&83.5&\textbf{83.7}&67.7&\textbf{86.4}&\textbf{85.2}&\underline{72.8}&\textbf{72.9}&86.7&84.7&\textbf{69.4}&64.9&\underline{86.3}&\underline{84.6} \\
\hspace{4mm}-pos+neg@dynamic
&70.6&69.7&85.1&\textbf{83.8}&68.8&66.0&\textbf{86.7}&\textbf{84.7}&\underline{75.9}&\textbf{74.7}&85.9&84.9&70.8&72.6&86.9&\underline{85.2}&67.7&\underline{65.6}&85.7&84.4 \\

\midrule
\multicolumn{9}{l}{\textit{MR$^2$ (cross-domain)}} \\

wo-RAG
&62.8&\underline{57.7}&68.4&\underline{60.3}&58.2&\textbf{55.5}&68.9&\textbf{65.0}&53.8&55.7&63.9&65.4&66.9&\textbf{57.8}&72.1&\underline{62.2}&66.6&\textbf{54.8}&74.0&\underline{68.2} \\

Text@full
&55.8&51.8&71.2&\underline{60.3}&40.5&50.9&63.9&61.8&59.4&54.6&60.5&63.9&59.1&56.9&72.7&\textbf{63.0}&52.8&52.7&73.6&66.8 \\

Text+Image@full
&66.3&51.2&\underline{73.2}&53.5&57.8&49.0&\textbf{73.2}&62.5&63.9&\textbf{58.8}&65.3&\textbf{66.7}&\textbf{69.5}&56.4&\textbf{73.9}&57.6&\textbf{67.3}&48.6&\textbf{75.1}&64.7 \\

\hline
\textbf{ExDR}$_\textit{evidence}$& & & & & & & & & & & & & & & & & & & &
 \\

\hspace{4mm}-single@full
&63.5&57.6&\textbf{73.6}&57.4&\underline{60.8}&50.6&69.1&60.5&66.6&42.3&\underline{65.2}&\underline{66.1}&68.1&55.0&73.1&59.3&65.4&52.1&72.2&64.5 \\

\hspace{4mm}-single@dynamic
&63.9&\textbf{57.8}&68.3&59.7&\textbf{61.0}&\underline{51.4}&70.3&62.9&62.3&57.1&64.7&65.8&68.4&56.8&72.2&59.9&65.7&\underline{54.2}&\underline{74.4}&\textbf{68.3} \\

\hspace{4mm}-pos+neg@full
&\textbf{67.6}&53.4&73.1&54.2&59.2&50.5&\underline{72.2}&61.7&\textbf{72.0}&44.0&\textbf{68.6}&\textbf{66.7}&\underline{68.6}&55.0&73.1&57.6&\underline{67.1}&51.3&73.4&64.1 \\

\hspace{4mm}-pos+neg@dynamic
&\underline{66.9}&53.3&70.9&\textbf{60.7}&59.0&50.4&\textbf{73.2}&\underline{64.2}&\underline{69.0}&\underline{57.6}&64.0&65.5&68.5&\underline{57.1}&\underline{73.2}&60.5&66.5&\textbf{54.8}&\underline{74.4}&68.1 \\

\bottomrule
\end{tabular}

\end{adjustbox}

\label{tab:method-results}
\end{table*}

\subsection{Datasets and Baselines}

\subsubsection{Datasets}
We evaluate our method on two Multimodal Fake News Detection benchmarks: \texttt{AMG}~\cite{guo2025each} and \texttt{MR$^2$}~\cite{hu2023mr2}, and conduct both in-domain and cross-domain evaluations according to the source of the retrieval corpus and the fine-tuning setting.
For the in-domain setting, we conduct experiments on \texttt{AMG}, the first dataset designed for attributing multimodal fake news with multi-granularity, covering five types of deception labels.
It contains a total of 5,022 samples, which are split into 3,532 for training, 517 for validation, and 973 for testing. Specifically, we use the training set $\mathcal{D}_{\text{train}}$ to construct the retrieval corpus $\mathcal{D}_{\text{corpus}}$. The validation set $\mathcal{D}_{\text{val}}$ is used for threshold selection, and the test set $\mathcal{D}_{\text{test}}$ is used for final evaluation.
For the cross-domain setting, we adopt \texttt{MR$^2$}, a large-scale multimodal and multilingual rumor detection dataset composed of image–text pairs. The dataset contains 14,700 samples in both Chinese and English. In this setting, the retrieval corpus and the fine-tuned model trained on \texttt{AMG} are directly reused without further adaptation, thereby introducing a realistic cross-domain retrieval scenario. In our experiments, we only use the English subset for the test set $\mathcal{D}_{\text{test}}$, which consists of 3,295 samples, to evaluate the cross-dataset generalization ability of our method.

\subsubsection{Baselines}
To ensure a fair and comprehensive comparison, we implemented several RAG-based baselines under a unified experimental setup. These baselines differ along two key components: the retrieval trigger and the evidence retrieval strategy.

For the \textit{retrieval trigger} module, we evaluate three strategies: (1) \textbf{FL-RAG}~\cite{ram2023context}, which activates retrieval when the input sequence exceeds a predefined token length; (2) \textbf{FLARE}~\cite{jiang2023active}, which triggers retrieval when any generated token has a probability below a threshold; and (3) \textbf{DRAGIN}~\cite{su2024dragin}, which considers the importance and uncertainty of generated tokens based on attention weights.

For the \textit{evidence retrieval} module, we compare three standard methods: (1) \textbf{wo-RAG}, using the vanilla model without any retrieval; (2) \textbf{Text@full}, employing BM25 to retrieve samples based on textual similarity; and (3) \textbf{Image-Text@full}~\cite{tang2024leveraging}, which computes retrieval scores using combined image-text features.

To evaluate the overall performance on the MFND task, we conduct end-to-end experiments and compare several traditional MFND models, including \textbf{MCAN}~\cite{wu-mcan}, \textbf{CAFE}~\cite{62chencafe}, \textbf{BMR}~\cite{yingbmr}, \textbf{CLIP}~\cite{radfordclip}, and \textbf{MGCA}~\cite{guo2025each}, all trained on the same dataset for fair comparison.

\vspace{-1mm}
\subsection{Evaluation Metrics}
We propose two new metrics to measure the effectiveness of dynamic trigger: \textit{Retrieval Identification Rate} and \textit{Retrieval Efficiency}. These metrics quantify (i) whether the model correctly identifies samples that require retrieval and (ii) whether retrieval leads to meaningful performance gains.

\noindent\textbf{Retrieval Identification Rate (RI)} measures the proportion of triggered samples that were misclassified before retrieval, reflecting triggering accuracy:
    \begin{equation}
        \text{RI} = \frac{N_{\text{err-classified}}}{N_{\text{retri.}}},
    \end{equation}
where $N_{\text{retri.}}$ denotes the number of triggered samples, and $N_{\text{err-classified}}$ is the subset of those that were incorrectly classified prior to retrieval. A higher RI indicates that the trigger more precisely identifies cases where retrieval is genuinely needed.
    
\noindent\textbf{Retrieval Efficiency (RE)} evaluates the cost-benefit trade-off of the dynamic trigger. It measures how much of the potential accuracy gain from full retrieval is achieved, normalized by retrieval cost:
\begin{equation}
\label{eq:re}
    \text{RE} = \frac{N_{\text{dyn}} - N_{\text{no}}}{N_{\text{full}} - N_{\text{no}}} \times \frac{N_{\text{total}}}{N_{\text{retrieved}}},
\end{equation}
where $N_{\text{dyn}}, N_{\text{full}},$ and $N_{\text{no}}$ are the counts of correct predictions with dynamic, full, and no retrieval, respectively. A higher RE indicates a more efficient strategy.

In most cases, dynamic and full retrieval exhibit consistent performance trends relative to the no-retrieval baseline, with the result that both tend to be either advantageous or disadvantageous. Accordingly, we report the standard RE value without additional symbols. However, in cases where one or both strategies underperform compared to the no-retrieval baseline, i.e., $N_{\text{dyn}} < N_{\text{no}}$ or $N_{\text{full}} < N_{\text{no}}$, we use special annotations to highlight their relative performance: a `$+$' signifies that dynamic retrieval is superior to full retrieval ($N_{\text{dyn}} > N_{\text{full}}$), while a `$-$' indicates the opposite.

For evidence retrieval methods and end-to-end MFND experiments, we adopt \textbf{Accuracy} and \textbf{F1 Score} as standard metrics to evaluate the fundamental classification performance in the fake news detection task.

\subsection{Experimental Settings}
We conduct experiments on \textbf{Qwen2.5-VL-Instruct} (7B, 32B)~\cite{bai2025qwen2}, \textbf{InternVL2.5} (8B, 26B)~\cite{chen2024expanding}, and \textbf{LLaVA-1.6-Mistral} (7B)~\cite{liu2024llavanext}. For Named Entity Recognition model, we use \texttt{bert-base-NER}, and for image encoding, we use \texttt{clip-vit-base-patch32}. The hyperparameter $K$ is set to 10. During threshold search, we employ $N_{\text{iter}} = 100$ samples for global exploration and select the top 5 for subsequent local refinement.
 For all retrieval triggering baseline models, we identify the optimal threshold on the validation set $\mathcal{D}_{\text{val}}$ and fix it for subsequent experiments. We adopt the LoRA technique~\cite{hu2022lora} to construct fine-tuned models for comparison, using the AMG training set $\mathcal{D}_{\text{train}}$. We use a LoRA rank of 8, a batch size of 4, and a learning rate of 1e-4, and run for 5 epochs. The implementation is executed on 2 A800 GPUs with 80GB of memory.

\subsection{Main Results}

We conducted experiments on the AMG and MR$^2$ dataset,  obtaining the following conclusions:

\noindent \textbf{(1) \textit{Multi-level confidence estimation enables more accurate and efficient retrieval triggering.}} 
As shown in Table~\ref{tab:trigger-results}, \textbf{ExDR}\textsubscript{trigger}, which dynamically performs retrieval based on multi-level confidence, consistently outperforms baselines in both Retrieval Indication (RI) and Retrieval Effectiveness (RE) metrics. This demonstrates its strong ability to decide when external evidence is necessary. The advantage is particularly pronounced on models like InternVL2.5-8B and LLaVA-1.6-Mistral-7B, where our trigger achieves significantly better performance than competing methods. 
Moreover,  on the fine-tuned InternVL2.5-26b model, both the fixed-length \textit{FL-RAG} and the token-based \textit{FLARE} methods fail to trigger any retrieval. In contrast, only the attention-weighted approach DRAGIN and our ExDR$_{trigger}$ successfully activate retrieval, with our method achieving overall superior performance. 
Furthermore, in the cross-domain evaluation on the MR$^2$ dataset, our trigger module also demonstrates strong robustness.

\noindent \textbf{(2) \textit{Entity-enriched multimodal hybrid indexing provides more discriminative cues for evidence retrieval.}}
When comparing single-sample retrieval methods, our entity-enriched indexing demonstrates a significant advantage over traditional multimodal indexing. As shown in Table~\ref{tab:method-results}, both \textit{ExDR$_{single@full}$} and \textit{Text+Image@full} retrieve one evidence sample, yet our method often achieves substantially better results. 
For instance, our \textit{ExDR$_{single@full}$} achieves \text{71.5\%} accuracy on the AMG dataset with vanilla LLaVA-1.6-7B, substantially outperforming the \textit{Text+Image@full} by \text{13.4\%}, despite both retrieving a single evidence sample. This advantage holds even in the challenging cross-domain MR$^2$ evaluation, where our method achieves \text{66.6\%} accuracy against the baseline's \text{63.9\%}. This consistent outperformance indicates that our explanation-guided entity indexing provides more discriminative cues than generic image-text embeddings, enabling the selection of more informative reference samples.

\textbf{(3) \textit{Contrastive evidence retrieval provides more informative context than single-polarity matching.}} While our entity-enriched hybrid indexing already achieves strong performance with single-sample retrieval, employing positive-negative contrastive evidence based on fine-grained deception labels yields additional improvements across multiple settings. 
This enhancement is most pronounced on vanilla LLaVA-1.6-Mistral-7B, where using positive-negative retrieval elevates performance from 71.5\% to 83.7\%, achieving the highest accuracy among all methods.  
Beyond in-domain performance, it also demonstrates particular value in cross-domain scenarios. For example, on MR$^2$ evaluation with vanilla LLaVA-1.6, it achieves 72.0\% accuracy compared to 66.6\% for single retrieval, showcasing superior generalization capability. 
These results suggest that contrastive evidence enables the selection of meaningful contrasts, allowing models to distinguish inconsistencies.

\begin{table}[t]
\small
\centering
\caption{Comparison results of Multimodal Fake News Detection on the AMG dataset. ExDR represents the average performance across five fine-tuned LVLMs.}
\label{tab:mfnd_result}
\setlength{\tabcolsep}{12pt} 
\begin{tabular}{lcc}
\toprule
\multirow{2}{*}{\textbf{MFND Method}} & \multicolumn{2}{c}{\texttt{AMG}} \\
 & \textit{ACC} & \textit{F1} \\
\midrule
CLIP & 78.1 & 78.1 \\
CAFE & 76.7 & 76.3 \\
MCAN & 77.4 & 76.9 \\
BMR  & 80.8 & 80.6 \\
MGCA & 83.2 & 83.1 \\
\midrule
\textbf{ExDR} & \textbf{86.1} & \textbf{84.6} \\
\bottomrule
\end{tabular}
\end{table}

\begin{figure*}
    \centering
    \includegraphics[width=1\linewidth]{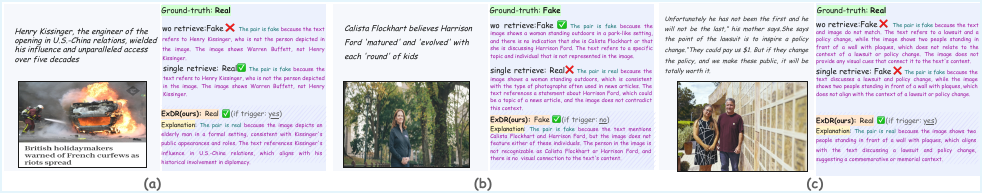}
    \caption{
Case study illustrating the effectiveness of ExDR across three representative scenarios.
}
    \label{fig:case-study}
\end{figure*}

\textbf{(4) \textit{ExDR demonstrates significant advantages over traditional models on the MFND task.}} We evaluate the end-to-end performance of the task, as shown in Table~\ref{tab:mfnd_result}. Experimental results indicate that our model consistently outperforms the baselines. Compared to the strong baseline MGCA, our model achieves improvements of 2.9\% in ACC and 1.5\% in F1, demonstrating the effectiveness of our approach.

\begin{table}[t!]
\centering
\caption{Ablations of core components in \textbf{ExDR}. Results are averaged over three untuned LVLMs: Qwen2.5-VL-7B, LLaVA-1.6-Mistral-7B, and InternVL2.5-8B.}

\setlength{\tabcolsep}{3pt}
\resizebox{\linewidth}{!}{
\begin{tabular}{c@{\hspace{2em}}c}  

\begin{tabular}{@{}l|cc|cc@{}}
\multicolumn{5}{c}{\textbf{(a) Retrieval Trigger Component}} \\
\toprule

\multirow{2}{*}{\textbf{Method}} & \multicolumn{2}{c|}{\texttt{AMG}} & \multicolumn{2}{c}{\texttt{MR$^2$}} \\
                                 & \textit{RI} & \textit{RE} & \textit{RI} & \textit{RE} \\
\midrule
\textbf{ExDR}$_\textit{trigger}$ & \textbf{49.0} & \textbf{1.70}  & \textbf{46.1} & 1.75  \\
w/o $\tau_\text{label}$ & 45.9 & 1.46  & 44.7 & 1.06 \\
w/o $\tau_\text{tok}$ & 43.3 & 1.36  & 43.4 & \textbf{1.82}  \\
w/o $\tau_\text{sent}$ & 47.9 & 1.61  & 45.7 & 1.20 \\
\midrule

\end{tabular}
&
\begin{tabular}{@{}l|cc|cc@{}}
\multicolumn{5}{c}{\textbf{(b) Evidence Retrieval Component}} \\
\toprule
\multirow{2}{*}{\textbf{Method}} & \multicolumn{2}{c|}{\texttt{AMG}} & \multicolumn{2}{c}{\texttt{MR$^2$}} \\
                                 & \textit{ACC} & \textit{F1} & \textit{ACC} & \textit{F1} \\
\midrule
\textbf{ExDR}$_\textit{evidence}$ & \textbf{74.9} & \textbf{68.1} & \textbf{66.3} & 49.3 \\
w/o entity feat. & 74.2 & 66.8 & 65.0 & 49.3 \\
w/o label det. & 71.0 & 64.7 & 63.6 & \textbf{50.2} \\
w/o both & 66.5 & 64.0 & 64.1 & 48.3 \\
\midrule

\end{tabular}
\end{tabular}
}

\label{tab:ablation}
\end{table}

\subsection{Ablation Study}
We conduct ablation studies to validate each component of our framework, with detailed results presented in Table~\ref{tab:ablation}.
The analysis of our retrieval trigger in Table~\ref{tab:ablation}(a) shows that the full ExDR$_{trigger}$, with all three thresholds, achieves the best overall performance. Removing any single dimension degrades performance, with token-level support $\tau_{\text{tok}}$ having the greatest impact.

For the evidence retrieval module, Table~\ref{tab:ablation}(b) shows that both the entity-enriched multimodal hybrid indexing and the label detection mechanism are critical.  Disabling label detection (\textit{w/o label det.}) causes the largest drop, highlighting the importance of label-voted contrastive retrieval. Likewise, removing entity features (\textit{w/o entity feat.}) degrades performance, confirming the utility of entity-aware signals for more targeted retrieval.  When both components are removed (\textit{w/o both}), the performance drops further, demonstrating their complementary roles in enhancing retrieval quality.

These ablation studies collectively validate the design choices in our ExDR, showing that each component makes a meaningful contribution to the overall performance.

\subsection{Case Study}
Figure~\ref{fig:case-study} presents three illustrative cases that demonstrate the key advantages of our ExDR framework.
The first scenario, depicted in Figure 2a,  involves a hard sample that the base model misclassifies. Our dynamic trigger correctly identifies this low confidence and initiates retrieval, which successfully corrects the prediction.
In the second case (b), the base model correctly identifies a simple sample, but naive retrieval introduces noise, leading to a misclassification. In contrast, our trigger correctly refrains from retrieval, preserving the original correct prediction and enhancing efficiency.
Finally, the third example (c) presents a complex case where both the base model and simple retrieval fail. Our ExDR combines an accurate trigger with explanation-guided retrieval, providing the targeted evidence required for a correct classification.

 \subsection{Quantitative Analysis}
We conduct a quantitative analysis by: (1) comparing different threshold search strategies to validate our two-stage approach; (2) varying the number of required confidence conditions to assess the benefit of combining three measures; and (3) analyzing the importance of each confidence measure in the final triggering decisions.

\subsubsection{Threshold Optimization Algorithm Analysis}

We compare three threshold search strategies: Grid search, Bayesian search, and our proposed two-stage hybrid search. 
As shown in Table~\ref{tab:threshold_strategy}, we observe that for models with smaller parameter sizes, Grid search and Bayesian search yield similar accuracy, while our hybrid approach with local optimization achieves significantly better performance. However, for larger models, the choice of threshold optimization strategy has less impact on performance, with all three methods producing comparable results. This suggests that larger models are less sensitive to threshold selection.

\begin{table}[htb]
\centering
\caption{Accuracy (\%) of different threshold search strategies on the AMG and MR$^2$ datasets. All results are obtained under the vanilla setting of LVLMs.}
\label{tab:threshold_strategy}

\setlength{\tabcolsep}{5pt}
\small
\begin{tabular}{lccccc}
\toprule
\textbf{Methods} &
\multicolumn{2}{c}{\textbf{Qwen-2.5-VL}} &
\multicolumn{2}{c}{\textbf{InternVL2.5}} &
\textbf{LLaVA-1.6} \\
\cmidrule(lr){2-3}\cmidrule(lr){4-5}\cmidrule(lr){6-6}
& \textbf{7B} & \textbf{32B} & \textbf{8B} & \textbf{26B} & \textbf{7B} \\
\midrule

\multicolumn{6}{l}{\textit{\textbf{AMG}}} \\
Grid Search    & 66.2 & 70.7 & 70.5 & 67.0 & 68.4 \\
Bayesian       & 66.7 & \textbf{71.7} & \textbf{71.0} & 66.8 & 67.8 \\
Hybrid (ours)  & \textbf{70.6} & 70.8 & 68.8 & \textbf{67.7} & \textbf{75.9} \\
\midrule
\multicolumn{6}{l}{\textit{\textbf{MR$^2$}}} \\
Grid Search    & 63.7 & 68.0 & \textbf{60.9} & 65.9 & 62.6 \\
Bayesian       & 63.7 & 68.1 & 60.7 & \textbf{66.5} & 62.4 \\
Hybrid (ours)  & \textbf{66.9} & \textbf{68.5} & 59.0 & \textbf{66.5} & \textbf{69.0} \\
\bottomrule
\end{tabular}
\end{table}

\subsubsection{Number of Required Conditions Analysis}

\begin{table}[t]
\centering
\caption{Analysis of threshold configurations for our two-stage hybrid search strategy across the AMG and MR$^2$ datasets, reporting accuracy (ACC) and retrieval trigger ratio for different numbers of thresholds (n).}
\begin{adjustbox}{width=0.49\textwidth}
\begin{tabular}{l|cc|cc|cc|cc|cc}
\toprule
\textbf{Methods} 
& \multicolumn{2}{c}{\textbf{Qwen-2.5-VL}} & \multicolumn{2}{c}{\textbf{InternVL2.5}} & \multicolumn{2}{c}{\textbf{LLaVA-1.6}} & \multicolumn{2}{c}{\textbf{Qwen-2.5-VL}} & \multicolumn{2}{c}{\textbf{InternVL2.5}} \\
& \multicolumn{2}{c}{(7B)} & \multicolumn{2}{c}{(8B)} & \multicolumn{2}{c}{(7B)} & \multicolumn{2}{c}{(32B)} & \multicolumn{2}{c}{(26B)} \\
\cmidrule(lr){2-3} \cmidrule(lr){4-5} \cmidrule(lr){6-7} \cmidrule(lr){8-9} \cmidrule(lr){10-11}
& \textit{ACC} & \textit{ratio} & \textit{ACC} & \textit{ratio} & \textit{ACC} & \textit{ratio} & \textit{ACC} & \textit{ratio} & \textit{ACC} & \textit{ratio}\\
\midrule
\midrule

\multicolumn{11}{l}{\textit{\textbf{AMG}}} \\

n=1
& 69.0 & 67.4 & 71.2 & 99.7 & 68.2 & 33.4 &72.3&40.7&68.2&39.0\\

n=2
& 67.0 & 30.9 & 71.0 & 97.7 & 68.1 & 35.4 &72.4&73.1&67.3&21.0\\

n=3 (ours)
& \text{70.6} & \text{67.4} & \text{68.8} & \text{59.2} & \text{75.9} & \text{27.2} &\text{70.8}&\text{54.6}&\text{67.7}&\text{10.8}\\

\midrule
\multicolumn{11}{l}{\textit{\textbf{MR$^2$}}} \\

n=1
& 63.9 & 86.1 & 60.8 & 99.8 & 62.4 & 54.0 &67.8&46.8&66.8&23.8\\

n=2
& 64.2 & 76.2 & 62.8 & 99.7 & 62.4 & 53.7 &68.1&20.7&66.3&8.9\\

n=3 (ours)
& \text{66.9} & \text{75.8} & \text{59.0} & \text{77.5} & \text{69.0} & \text{38.8} &\text{68.5}&\text{45.8}&\text{66.5}&\text{9.5}\\

\bottomrule
\end{tabular}
\end{adjustbox}

\label{tab:amg-mr2-full}
\end{table}

We examine the effect of varying the number of required threshold conditions in our two-stage hybrid search. Specifically, we test three settings: 

(1) $n=3$: all three thresholds must be satisfied (our method);

(2) $n=2$: any two thresholds must be satisfied; 

(3) $n=1$: only one threshold needs to be satisfied. 

As shown in Table~\ref{tab:amg-mr2-full}, requiring more conditions consistently lowers the trigger ratio while maintaining, or even improving, recognition accuracy. This suggests that our confidence estimation can suppress unnecessary retrievals without hurting predictive performance, supporting the design choice of enforcing all three confidence measures for optimal results.

\subsubsection{Importance Analysis of Confidence Measures}

\begin{figure}[ht!]
    \centering
    \begin{minipage}{0.4\columnwidth}
        \centering
        \includegraphics[width=\linewidth]{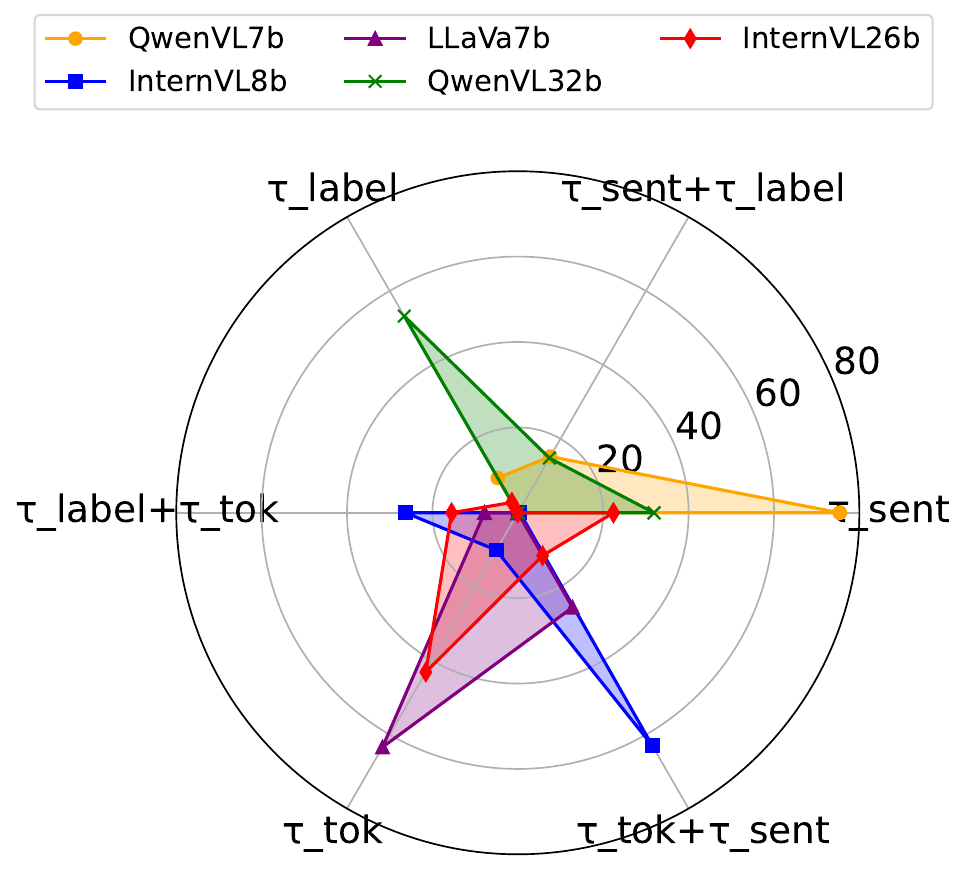}
        \subcaption{AMG - Vanilla}
        \label{fig:amg_org}
    \end{minipage}
    \hfill
    \begin{minipage}{0.4\columnwidth}
        \centering
        \includegraphics[width=\linewidth]{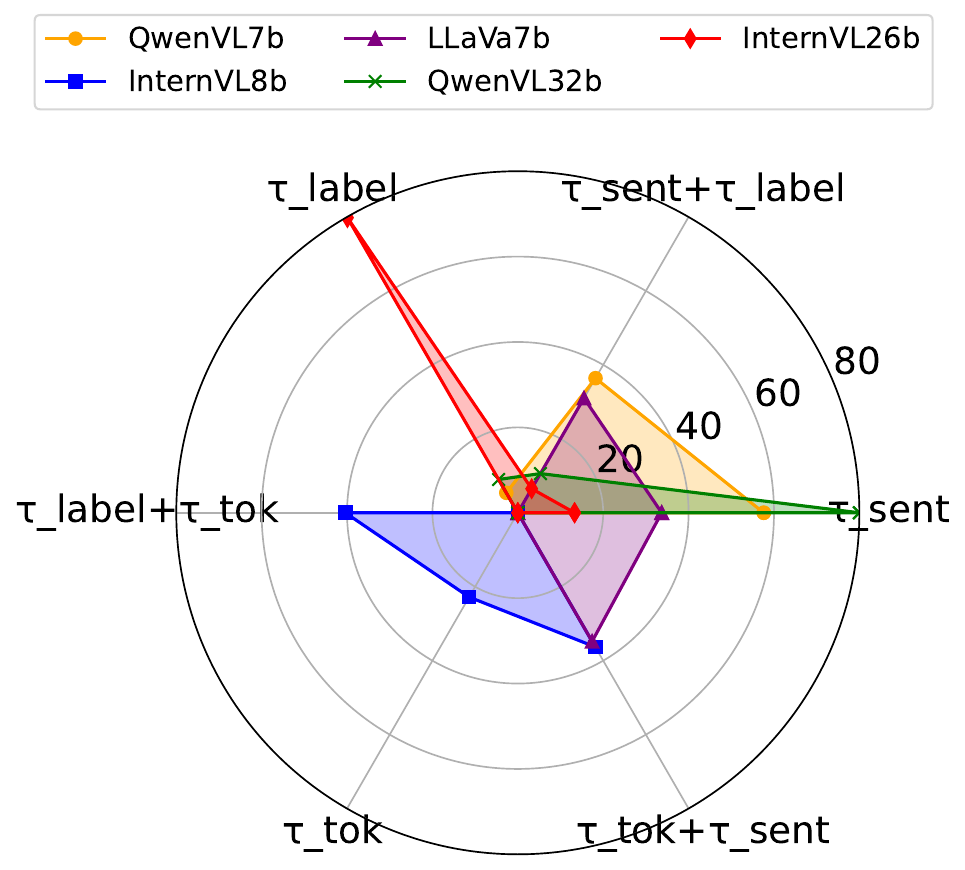}
        \subcaption{MR$^2$ - Vanilla}
        \label{fig:mr2_org}
    \end{minipage}
    
    \vspace{0.5em}
    
    \begin{minipage}{0.4\columnwidth}
        \centering
        \includegraphics[width=\linewidth]{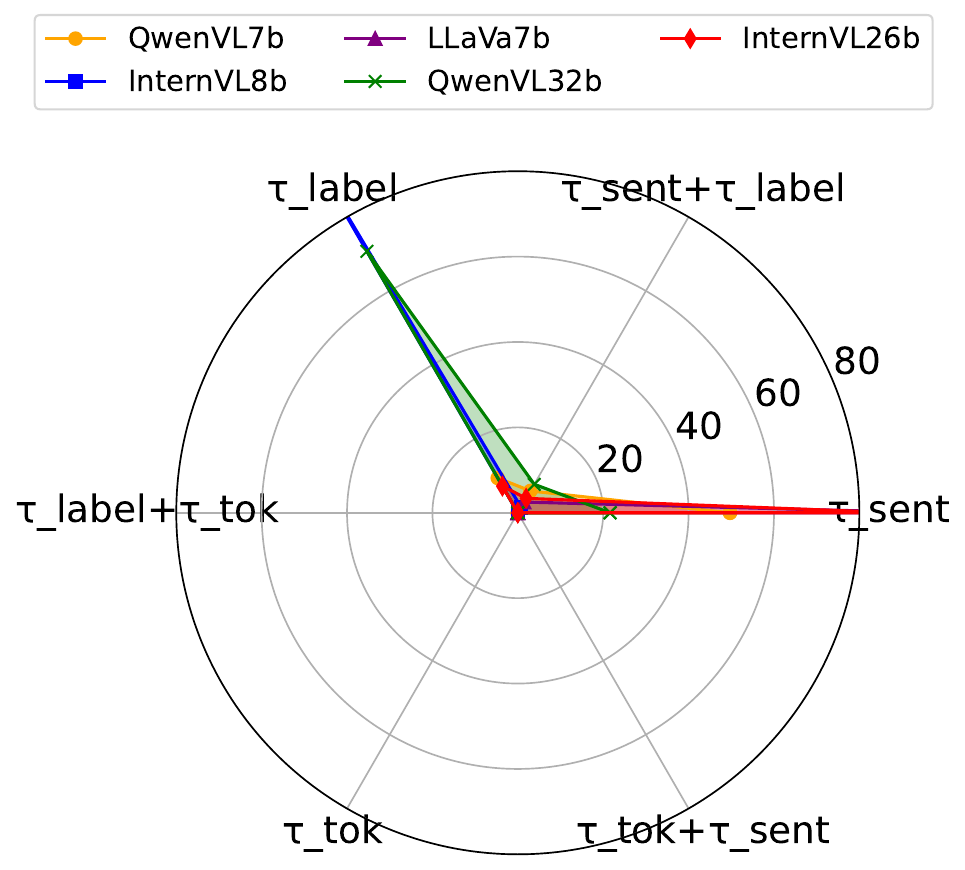}
        \subcaption{ AMG - Fine-tuned}
        \label{fig:amg_ft}
    \end{minipage}
    \hfill
    \begin{minipage}{0.4\columnwidth}
        \centering
        \includegraphics[width=\linewidth]{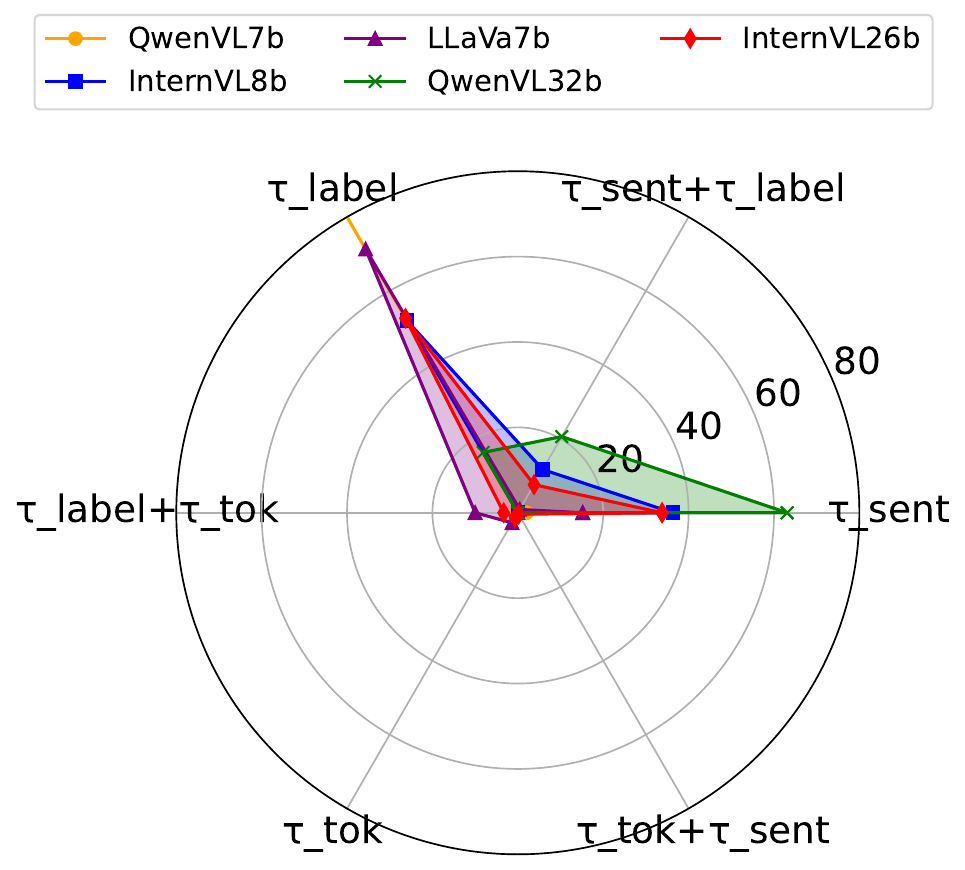}
        \subcaption{ MR$^2$ - Fine-tuned}
        \label{fig:mr2_ft}
    \end{minipage}
    \caption{Contribution analysis of different confidence measures to retrieval triggering.}
    \label{fig:importance_analysis}
\end{figure}
We analyze which confidence measures contribute most to successful retrieval triggering. Using the $n=1$ and $n=2$ threshold condition, we examine the contribution of different threshold combinations to the final retrieval decisions. Figure~\ref{fig:importance_analysis} presents the results.

For the vanilla models, as shown in Figures~\ref{fig:importance_analysis}(a) and \ref{fig:importance_analysis}(b), all three thresholds play important roles, validating the necessity of using all three confidence measures. However, for fine-tuned models, depicted in Figures~\ref{fig:importance_analysis}(c) and \ref{fig:importance_analysis}(d), supervised fine-tuning tends to homogenize token-level probability distributions, which diminishes the discriminative power of token-level signals. As a result, token-level support contributes minimally, indicating a reduced reliance on this particular confidence measure.

\section{Conclusion}
In this paper, we propose \textbf{ExDR}, a novel Explanation-driven Dynamic Retrieval framework specifically designed to address the unique challenges of multimodal fake news detection. Our framework focuses on three key aspects of retrieval augmentation: when to retrieve, how to retrieve, and what to retrieve. By introducing a dynamic triggering mechanism that evaluates the confidence of model-generated explanations and an evidence retrieval strategy driven by entity-enriched indexing and contrastive evidence, \text{ExDR} improves both the efficiency and precision of the retrieval process. Extensive experiments on the AMG and MR$^2$ datasets demonstrate that our approach consistently outperforms existing methods in both fake news detection accuracy and retrieval effectiveness, highlighting its strong potential for building more reliable and robust misinformation detection systems.


\appendix

\begin{table*}[htb]
\small
\centering
\caption{Definitions of fine-grained deception labels and their explanations.}
\renewcommand{\arraystretch}{1}  
\setlength{\tabcolsep}{6pt}         
\rowcolors{2}{gray!10}{white}       
\begin{tabular}{>{\bfseries}l m{12.5cm}}
\toprule
Fine-grained Deception Label & \textbf{Explanation} \\
\midrule
Real News & The image and text are consistent and factually aligned, representing authentic news. \\
Image Fabrication & The image is likely manipulated or artificially generated. This includes advanced deepfake techniques, simple edits like splicing or Photoshopping, and fabricated visuals imitating official websites or social media posts. \\
Entity Inconsistency & The key entities (e.g., people, places, organizations) mentioned in the text do not match those depicted in the image, indicating a misalignment between modalities. \\
Event Inconsistency & Although the image and text refer to the same entities, they describe different events. The image may be genuine, but the text misrepresents or over-interprets its context. \\
Time or Space Inconsistency & The image or video is real but misrepresented in time or location---for example, showing past or distant events as if they were current or local. \\
Ineffective Visual Information & The image provides no evidential support for the accompanying text. It typically consists of text-only visuals or irrelevant content, lacking concrete proof such as on-site or event-related imagery. \\
\bottomrule
\end{tabular}

\label{tab:fine_label}
\end{table*}

\section{Fine-grained Deception Labels}
\label{appendix:deception_label}
In addition to the binary labels \textsc{Real} and \textsc{Fake}, the AMG dataset further annotates each fake news sample with a fine-grained deception label. Specifically, the fine-grained labels include \textit{Image Fabrication}, \textit{Entity Inconsistency}, \textit{Event Inconsistency}, and \textit{Time or Space Inconsistency}. The extended deception labels $\tilde{y}_i$ and their corresponding explanations are presented in Table~\ref{tab:fine_label}.

\section{Human Evaluation of LLM Explanations}
\label{appendix:human_verify}
We conduct a human evaluation of LLM-generated explanations. Three annotators independently assess 50 randomly sampled instances, given the multimodal input, its label, and the explanation generated by GPT-4o. They rate (0/1/2): (i) \textbf{Label--Explanation Consistency} (\textsc{Consis.}), whether the explanation supports the label; (ii) \textbf{Multimodal Faithfulness} (\textsc{MM-Faith.}), whether it is grounded in the image and text without hallucinated entities or unsupported claims; and (iii) \textbf{Evidence Usefulness} (\textsc{Useful.}), whether it provides specific, actionable clues rather than generic statements. We report mean scores and inter-annotator agreement (Fleiss' $\kappa$) with majority-vote aggregation. We observe substantial agreement ($\kappa{=}0.58$), and the averaged scores are \textsc{Consis.}=$\mathbf{1.94}$, \textsc{MM-Faith.}=$\mathbf{1.75}$, and \textsc{Useful.}=$\mathbf{1.89}$. These results suggest the explanations are generally reliable and useful for retrieval.

\section{Prompt Templates}
\label{sec:appendix_prompting}
For the final prediction, we adopt two prompting strategies corresponding to the two modes. \textbf{Prompt} \textsc{II} is used in the non-retrieval setting, i.e., for baselines without retrieval or when the dynamic trigger deems external evidence unnecessary. When retrieval is activated, we use \textbf{Prompt} \textsc{III}, which prepends the retrieved positive ($\mathbf{x}^{+}$) and negative ($\mathbf{x}^{-}$) instances as few-shot in-context examples to guide a more informed and robust classification.

\section{Details for Token-Level Support Calculation}
\label{appendix:Lexicons}
The token-level support score, $\tau_{\text{tok}}$, quantifies the semantic alignment of the top-$K$ candidate tokens with the model's predicted label. This is determined via a hierarchical matching process.

\noindent\textbf{Step 1: Lexical Matching.} We compare each candidate token $t_i$ against  two predefined label-related lexicons:

\begin{itemize}[leftmargin=4mm]
    \item $\mathcal{L}_{\text{real}}$: \texttt{\{real, genuine, authentic, true, legitimate, realistic, legit, fact, accurate, related, likely, consistent, plausible\}}
    \item $\mathcal{L}_{\text{fake}}$: \texttt{\{fake, missing, false, fabric,  fict, un, mis, fraud, unrelated, fictional, inconsistent\}}
\end{itemize}

If a candidate token $t_i$ exactly matches any entry in either lexicon, it is directly assigned the corresponding label (\textsc{Real} or \textsc{Fake}).

\vspace{1mm}
\noindent\textbf{Step 2: Semantic Similarity Voting.} If $t_i$ is not found in either lexicon, we assess its alignment via semantic similarity. The token is inserted into a prompt template to form a query sentence: ``\texttt{This post is $t_i$.}'' and embedded using the \texttt{all-MiniLM-L6-v2} sentence encoder. It is then compared with reference sentence embeddings constructed from each lexicon:

\begin{itemize}[leftmargin=3mm]
    \item Real references: ``\textit{The post is \{word\} and factually correct.}'' for each word in $\mathcal{L}_{\text{real}}$
    \item Fake references: ``\textit{The post is \{word\} and contains misinformation.}'' for each word in $\mathcal{L}_{\text{fake}}$
\end{itemize}

We compute the mean cosine similarity between the query embedding and each reference group:

\begin{align}
\text{sim}_{\text{real}}(t_i) &= \frac{1}{|\mathcal{L}_{\text{real}}|} \sum_{w \in \mathcal{L}_{\text{real}}} \cos(\mathbf{e}_{\text{query}}, \mathbf{e}_{\text{real}, w}) \\
\text{sim}_{\text{fake}}(t_i) &= \frac{1}{|\mathcal{L}_{\text{fake}}|} \sum_{w \in \mathcal{L}_{\text{fake}}} \cos(\mathbf{e}_{\text{query}}, \mathbf{e}_{\text{fake}, w})
\end{align}
where $\mathbf{e}_{\text{query}}$ denotes the embedding of the query, $\mathbf{e}_{\text{real}, w}$ and $\mathbf{e}_{\text{fake}, w}$ denote the embeddings of the reference token $w$ in the real and fake groups, respectively, and $\cos(\cdot,\cdot)$ denotes cosine similarity.

The token is assigned the label with the higher similarity. This enables generalization to unseen but semantically aligned tokens.

\noindent\textbf{Step 3: Score Computation.} The final token-level support score is defined as the proportion of top-$K$ tokens whose assigned label matches the model's predicted label $y$:

\begin{equation}
\tau_{\text{tok}} = \frac{N_{\text{sup}}}{K},
\end{equation}
where $N_{\text{sup}}$ is the number of supporting tokens that agree with $y$ in either the lexical or semantic stage.

\begin{tcolorbox}[title=Prompt \textsc{II}: Fake News Detection without Retrieval, boxrule=0pt, left=1mm, right=1mm, top=1mm, bottom=1mm, fontupper=\small]
You are a knowledgeable and analytical fact-checking assistant. Your task is to determine whether a social text-image pair is fake.\\

Your response should be either \texttt{The pair is fake because \{explanation of your reasoning\}.} if the text and image present false, misleading, or manipulated content, or \texttt{The pair is real because \{explanation of your reasoning\}.} if the text and image are consistent and factually aligned.\\

Your explanation must be concise and clear, highlighting linguistic, visual, or contextual cues that support your conclusion.

\vspace{1mm}
\textbf{USER}: the image \textit{\textless image\textgreater} and the text \textit{\textless content \textgreater}.

\textbf{ASSISTANT}:
\end{tcolorbox}

\begin{tcolorbox}[title=Prompt \textsc{III}: Fake News Detection with Retrieval, boxrule=0pt, left=1mm, right=1mm, top=1mm, bottom=1mm, fontupper=\small]
You are a knowledgeable and analytical fact-checking assistant. Your task is to determine whether a social text-image pair is fake.\\

Your response should be either \texttt{The pair is fake because \{explanation of your reasoning\}.} if the text and image present false, misleading, or manipulated content, or \texttt{The pair is real because \{explanation of your reasoning\}.} if the text and image are consistent and factually aligned.\\

Your explanation should be concise and clear, highlighting any linguistic, visual, or contextual cues that support your conclusion.

\vspace{1mm}
\textbf{Refer to these examples:}

\textbf{USER}: the first image \textit{\textless positive\_image\textgreater} and the text \textit{\textless positive\_content\textgreater}.

\textbf{ASSISTANT}: The pair is real because \{positive\_explanation\}

\textbf{USER}: the second image \textit{\textless negative\_image\textgreater} and the text \textit{\textless negative\_content\textgreater}.
    
\textbf{ASSISTANT}: The pair is fake because \{negative\_explanation\}

\vspace{1mm}
\textbf{Now determine the following:}

\textbf{USER}: the third image \textit{[image]} and the text \textit{[content]}.

\textbf{ASSISTANT}:
\end{tcolorbox}


\bibliographystyle{ACM-Reference-Format}
\balance
\bibliography{sample-base}


\end{document}